\documentclass{article}

\usepackage[final]{neurips_2018}
\usepackage{comment}

\usepackage[utf8]{inputenc} 
\usepackage[T1]{fontenc}    
\usepackage{hyperref}       
\usepackage{url}            
\usepackage{booktabs}       
\usepackage{amsfonts}       
\usepackage{nicefrac}       
\usepackage{microtype}      

\usepackage{graphicx}
\usepackage{caption}
\usepackage{subcaption}
\usepackage{algorithm}
\usepackage{algorithmic}
\usepackage{relsize}
\usepackage{amsmath}
\usepackage{changes}
\usepackage{todonotes}
\usepackage{wrapfig}

\newcommand{\seq}{\!\!=\!\!}

\usepackage{hyperref}

\usepackage{lipsum}

\title{ Cooperative neural networks (CoNN): 
        Exploiting prior independence structure for improved classification}

\author{
   Harsh Shrivastava
   \thanks{Dept. of Comp. Sci. \& Eng. Georgia Institute of Technology
           Atlanta, GA 30332} \\
   Georgia Tech\\
   \texttt{\small hshrivastava3@gatech.edu} \\
	\And
   Eugene Bart
   \thanks{3333 Coyote Hill Rd, Palo Alto, CA,}\\
   PARC\\
   \texttt{\small bart@parc.com} \\
   \And 
   Bob Price \footnotemark[2]\\
   PARC\\
   \texttt{bprice@parc.com} \\
   \And
   Hanjun Dai \footnotemark[1]\\
   Georgia Tech\\
   \texttt{hanjundai@gatech.edu} \\
   \And
   Bo Dai \footnotemark[1]\\
   Georgia Tech\\
   \texttt{bodai@gatech.edu} \\
   \And
   Srinivas Aluru \footnotemark[1]\\
   Georgia Tech\\
   \texttt{aluru@cc.gatech.edu} \\
}

\begin{document}

\maketitle

\begin{abstract}
We propose a new approach, called cooperative neural networks (CoNN),
which uses a set of cooperatively trained neural networks 
to capture latent representations
that exploit prior given independence structure. 
The model is more flexible than traditional graphical models
based on exponential family distributions, 
but incorporates more domain specific prior structure than 
traditional deep networks or variational autoencoders. 
The framework is very general and can be used to exploit 
the independence structure of any graphical model.
We illustrate the technique by showing that we can transfer the 
independence structure of the popular Latent Dirichlet
Allocation (LDA) model to a cooperative neural network, CoNN-sLDA.
Empirical evaluation of CoNN-sLDA on supervised text classification tasks
demonstrates that the theoretical advantages of prior independence structure
can be realized in practice -
we demonstrate a 23\% reduction in error 
on the challenging MultiSent data set compared to state-of-the-art.
\end{abstract}

\section{Introduction}


Neural networks offer a low-bias solution 
for learning complex concepts
such as the linguistic knowledge required 
to separate documents into thematically related classes. 
However, neural networks typically start with a fairly generic structure, 
with each level comprising a number of functionally equivalent neurons connected to other layers by identical, repetitive connections. 
Any structure present in the problem domain 
must be learned from training examples and encoded as weights. 
In practice, some domain structure is often known ahead of time; in such cases, 
it is desirable to pre-design a network with this domain structure in mind. 
In this paper, we present an approach that allows incorporating 
certain kinds of independence structure into 
a new kind of neural learning machine.

The proposed approach is called ``Cooperative Neural Networks'' (CoNN). 
This approach works by constructing 
a set of neural networks, each trained to output an embedding of a probability distribution. 
The networks are iteratively 
updated so that each embedding is consistent
with the embeddings of the other networks and with the training data.
Like probabilistic graphical models, 
the representation is factored into components that are independent.
Unlike probabilistic graphical models, which are limited to tractable
conditional probability distributions (e.g., exponential family),
CoNNs can exploit powerful generic distributions represented by non-linear neural networks.
The resulting approach allows us to create models that
can exploit both known independence structure 
as well as the expressive powers of neural networks
to improve accuracy over competing approaches.

We illustrate the general approach of cooperative neural networks 
by showing how one can transfer the independence structure
from the popular Latent Dirichlet Allocation (LDA) model \cite{blei2003latent} to a set of 
cooperative neural networks.
We call the resultant model CoNN-sLDA.
Cooperative neural networks are different from feed forward networks
as they use back-propagation to enforce consistency 
across variables within the latent representation.
CoNN-sLDA improves over LDA as it admits more complex
distributions for document topics and better generalization over
word distributions. CoNN-sLDA is also better than 
a generic neural network classifier as the factored representation
forces a consistent latent feature representation that has a natural 
relationship between topics, words and documents.
We demonstrate empirically that the theoretical advantages of cooperative neural networks are realized in practice by showing that our CoNN-sLDA model 
beats both probabilistic and neural network-based state-of-the-art alternatives. We emphasize that
although our example is based on LDA, the CoNN approach is general and can be used with other graphical  models, as well as other sources of 
independence structure (for example, physics- or biology-based constraints).

\section{ Related Work }

Text classification has a long history beginning with the use of support vector machines on text features \cite{joachims98}. 
More sophisticated approaches integrated 
unsupervised feature generation and classification
in models such as sLDA
\cite{mcauliffe2008supervised, chong2009simultaneous} 
and discriminative LDA (discLDA) ~\cite{lacoste2009disclda}
and a maximum margin based combination \cite{zhu2009medlda}. 

One limitation of LDA-based models is that they pick topic distributions 
from a Dirichlet distribution and cannot represent the joint 
probability of topics in a document 
( i.e., hollywood celebrities, politics and business are all popular categories,
but politics and business appear together more 
often than their independent probabilities would predict). 
Models such as pachinko allocation \cite{li06} attempt to address this 
with complex tree structured priors. 
Another limitation of LDA stems from the fact that 
word topics and words themselves 
are selected from categorical distributions. These admit arbitrary empirical distributions over tokens, but don't generalize what they learn.
Learning about the topic for the token "happy" 
tells us nothing about the token "joyful".

There have been many generative deep learning models such as Deep Boltzmann Machines \cite{srivastava2013modeling}, NADE \cite{larochelle2012neural,zheng2016deep},  variational auto-encoders (VAEs) \cite{yang2017improved} and variations \cite{miao16}, GANs\cite{gan2015scalable}
and other deep generative networks 
\cite{tang2013learning, bengio2014deep, rezende2014stochastic,mnih2014neural}
which can capture complex joint distributions of words in documents 
and  surpass the performance of LDA.  
These techniques have proven to be good generative models. 
However, as purely generative models, they need a separate classifier to assign documents to classes. As a result, they are not trained end-to-end for the actual discriminative task that needs to be performed. Therefore, the resulting representation that is learned does not incorporate any problem-specific structure, leading to limited classification performance. 
Supervised convolutional networks have been applied to 
text classification \cite{kim14} but are limited to small fixed inputs
and still require significant data to get high accuracy.
Recurrent networks have also been used to handle open ended text \cite{dieng2016}.
A supervised approach for LDA with DNN was developed by \cite{chen2015end,chien2018deep} using end-to-end learning for LDA by using Mirror-Descent back propagation 
over a deep architecture called BP-sLDA. 
To achieve better classification, they have to increase the number of layers of their model, which results in higher model complexity, thereby limiting the capability of their model to scale. 
In summary, there are still significant challenges to creating
expressive, but efficiently trainable and computationally tractable models.

In the face of limited data, 
regularization techniques are an important way of trying to 
reduce overfitting in neural approaches.
The use of pretrained layers for networks is a key regularization strategy;
however, training industrial applications 
with domain specific language and tasks remains challenging. 
For instance, classification of field problem reports must 
handle content with arcane technical jargon, abbreviations and phrasing
and be able to output task specific categories.

Techniques such as L2 normalization of 
weights and random drop-out \cite{JMLR:v15:srivastava14a} 
of neurons during training are now widely used but provide little problem specific advantage. Bayesian neural networks with distributions have been proposed, but independent distributions over weights result in network weight means where the variance must be controlled fairly closely so that relative relationship of weights produces the desired computation.
Variational auto-encoders explicitly enable probability distributions
and can therefore be integrated over, 
but are still largely undifferentiated structure of identical units. 
They don't provide a lot of prior structure to assist with limited data.

Recently there has been work incorporating other kinds of 
domain inspired structure into networks such  Spatial transformer networks \cite{jaderberg2015}, capsule networks \cite{sabour2017dynamic}
and natural image priors \cite{hojjat2018}.

\section{Deriving Cooperative Neural Networks}

Application of our approach proceeds in several distinct steps.
First, we define the independence structure for the problem. 
In our supervised text classification example, 
we incorporate structure from latent dirichlet allocation (LDA)
by choosing to factor the distribution over document texts into 
document topic probabilities and word topic probabilities.
This structure naturally enforces the idea that 
there are topics that are common across all documents
and that documents express a mixture of these topics 
independently through word choices.
Second, a set of inference equations is derived from the independence structure. 
Next, the probability distributions involved in the variational approximation, 
as well as the inference equations, 
are mapped into a Hilbert space to reduce limitations on their functional form. 
Finally, these mapped Hilbert-space equations 
are approximated by a set of neural networks (one for each constraint), 
and inference in the Hilbert space is performed by iterating these networks. 
We call the combination of Cooperative Neural Networks and LDA as 
Cooperative Neural Network supervised Latent Dirichlet Allocation, or `CoNN-sLDA'.
These steps are elaborated in the following sections. 

\subsection{LDA model}

\begin{figure}
   
   \begin{subfigure}[t]{0.45\textwidth}
      \centering
      \includegraphics[width=\linewidth]{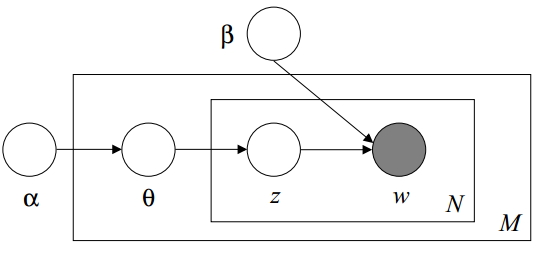}
      \caption{
        {
        LDA summarizes the content of each document $m$ in $M$ 
        as a topic distribution $\theta_m$. 
        Each word $w_{m,n}$ in $N_m$ has topic $z_{m,n}$ drawn from $\theta_m$.  } 
      }
      \label{basic_lda}
   \end{subfigure}%
   \hspace{0.09\textwidth}
   \begin{subfigure}[t]{0.45\textwidth}
      \centering
      \includegraphics[width=.5\linewidth]{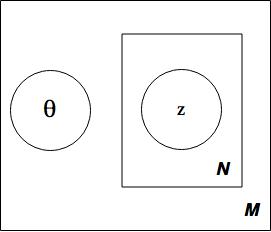}
      \caption{
         {
         Variational LDA approximates the posterior
         topic distribution $\theta_m$
         and word topic $z_{m,n}$ with independent distributions.}
      }
      \label{lda_pgm2}
   \end{subfigure}
  \caption{Plate models representing the original LDA and its approximation.}
\end{figure}

Here, we use the same notation and the same plate diagram (Figure~\ref{basic_lda}) 
as in the original LDA description \cite{blei2003latent}.
Let $K$ be the number of topics, 
$N$ be the number of words in a document, 
$V$ be the vocabulary size over the whole corpus, and 
$M$ be the number of documents in the corpus. 
Given the prior over topics $\alpha$ 
and topic word distributions $\beta$, 
the joint distribution over 
the latent topic structure $\theta$,
word topic assignments $\textbf{z}$,
and observed words in documents \textbf{w} 
is given by:
\begin{equation}\label{true}
   p( \theta, \textbf{z}, \textbf{w} | \alpha, \beta) 
      = p( \theta | \alpha ) 
        { \displaystyle \prod_{i=1}^{N} p( z_i | \theta ) p( w_i | z_i, \beta) }
\end{equation}

\subsection{Variational approximation to LDA}

Inference in LDA requires estimating the distribution over $\theta$ and $\textbf{z}$.
Using the Bayes rule, this posterior can be written as follows: 
\begin{equation}\label{Bayes}
   p( \theta, \textbf{z} | \textbf{w}, \alpha, \beta ) = 
             \frac{  p( \theta, \textbf{z}, \textbf{w} | \alpha, \beta ) }
                  {  p( \textbf{w} | \alpha, \beta ) }
\end{equation}

Unfortunately, directly marginalizing out $\theta$ in the original
model is intractable. 
Variational approximation of $p(\theta,\textbf{z})$ is a common work-around.
To perform variational approximation, 
we approximate this LDA posterior with the Probabilistic Graphical Model (PGM) shown in Figure~\ref{lda_pgm2}.
The joint distribution for the approximate PGM is given by:
\begin{equation}\label{e2}
   q( \theta, \textbf{z} ) = 
         q( \theta ) { \displaystyle \prod_{i=1}^{N} q_i( z_{i} )  }
\end{equation}

We want to tune the approximate distribution to resemble the true posterior as much as possible. 
To this end, we minimize the KL divergence 
between the two distributions. 
Alternatively, this can be seen as minimizing the variational free energy 
of the Mean-Field inference algorithm \cite{wainwright2008graphical}: 
\begin{equation}
   \underset{ \{q\} }{\rm min} 
         \left\{ D_{KL} (\;          q( \theta, \textbf{z} ) 
                            \;||\;   p( \theta, \textbf{z}  |  \textbf{w}, \alpha, \beta  )   \;)  \right\}
\end{equation}
To solve this minimization problem, we derive a set of fixed-point equations in Appendix(A). These fixed-point equations can be expressed as
\begin{equation}\label{up1}
\begin{split}
       \log{q(\theta)} 
               =  \log{p(\theta|\alpha)} 
                  +  &{\displaystyle \sum_{i=1}^{N} 
                                \int_{z_i} q_i(z_i) \log{p(z_i|\theta)}} 
                                     \;dz_{i} 
                  - 1
\end{split}
\end{equation}
\begin{equation}\label{up2}
      \log{q_i(z_{i})} =  \log{p(w_i|z_i, \beta)} 
                          + \int_{\theta} q(\theta) \log{p(z_i|\theta)} d\theta -1
\end{equation}
This set of equations is difficult to solve analytically. In addition, even if it was possible to solve them analytically, they are still subject to the limitations of the original graphical models, such as the need to use exponential family distributions and conjugate priors for tractability. 

Therefore, the next step in the proposed method is to map the probability distributions and the corresponding fixed-point equations into a Hilbert space, where some of these limitations can be relaxed. Section \ref{SeqGeneralEmbedding} gives a general overview of Hilbert space embeddings, and section \ref{SecLDAEmbedding} derives the corresponding equations for our model.

\subsection{Hilbert Space Embeddings of Distributions}
\label{SeqGeneralEmbedding}

We follow the notations and procedure defined in \cite{dai2016discriminative} 
for parameterizing Hilbert spaces. 
By definition, the Hilbert Space embeddings of probability distributions 
are mappings of these distributions into potentially \textit{infinite}
-dimensional feature spaces.
\cite{smola2007hilbert}. 
For any given distribution $p(X)$ and a feature map $\phi(x)$, 
the embedding $\mu_X:\mathcal{P}\rightarrow \mathcal{F}$ is defined as:
\begin{equation}\label{embed_eqn}
   \mu_X 
      \; := \;  E_X[\phi(X)] 
      \; = \;   \int_{\mathcal{X}} 
                          \phi(x) p(x) dx 
\end{equation}

For some choice of feature map $\phi$, 
the above embedding of distributions becomes injective \cite{sriperumbudur2008injective}. 
Therefore, any two distinct  distributions $p(X)$ and $q(X)$ 
are mapped to two distinct points in the feature space. 
We can treat the injective embedding $\mu_X$  
as a sufficient statistic of the corresponding probability density. 
In other words, $\mu_X$ preserves all the information of $p(X)$. 
Using $\mu_X$, we can uniquely recover $p(X)$ 
and any mathematical operation on $p(X)$ 
will have an equivalent operation on $\mu_X$. 
These properties lead to the following equivalence relations. 
We can compute a functional $f: \mathcal{P}\rightarrow \rm I\!R$ 
of the density $p(X)$ using only its embedding, 
\begin{equation}\label{hil1}
f(p(x)) = \tilde{f}(\mu_X)
\end{equation}
by defining $\tilde{f}: \mathcal{F} \rightarrow \rm I\!R$ 
as the operation on $\mu_X$ equivalent to $f$. 
Similarly, we can generalize this property to operators. 
An operator $\mathcal{T}:\mathcal{P} \rightarrow \rm I\!R^d$ 
applied to a density can also be equivalently carried out using its embedding,
\begin{equation}\label{hil2}
\mathcal{T}\circ p(x) = \tilde{\mathcal{T}} \circ \mu_X
\end{equation}
where $\tilde{\mathcal{T}}:\mathcal{F}\rightarrow \rm I\!R^d$ 
is again the corresponding equivalent operator applied to the embedding. 
In our derivations, we assume that there exists a feature space 
where the embeddings are injective 
and apply the above equivalence relations in subsequent sections. 


\begin{figure*}[t]  
  \begin{center}
  \includegraphics[width=\textwidth]{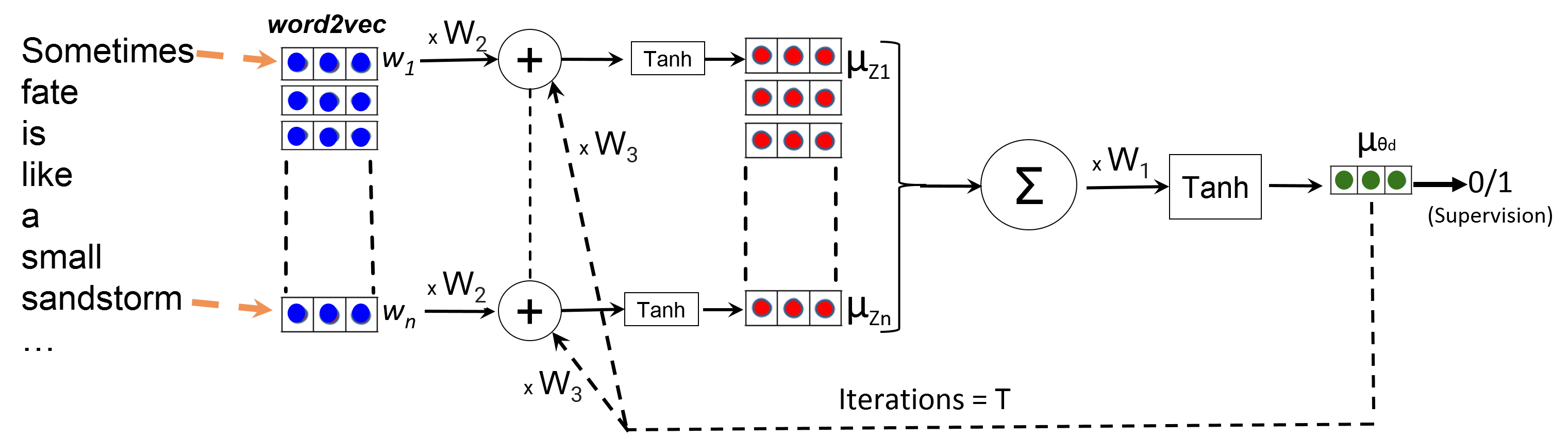}
  \caption{Visualization of the CoNN-sLDA architecture for a single document. For the $i$'th word, the latent topic variable is $z_i$. The embedding for the distribution $p(z_i)$ is $\mu_{z_i}$; these embeddings are shown as three-dimensional vectors for illustration. They are accumulated and passed through a non-linearity to obtain $\mu_\theta$, which is the embedding of $p(\theta)$, the distribution over the topics for the document. Thus, the embedding $\mu_\theta$ is determined (up to the non-linearity) by the average of the embeddings $\mu_{z_i}$, as in the original LDA model. Similarly, there is feedback from $\mu_\theta$ (which happens for $T$ iterations, see Alg\ref{Hilbert embedding}), so that $\mu_\theta$, in turn, influences $\mu_{z_i}$, again, as in the original LDA model.}
  \label{FigEmbeddingIllustration}
  \end{center}
  \vskip -0.1in
\end{figure*}

\subsection{Hilbert space embedding for LDA}
\label{SecLDAEmbedding}
We consider Hilbert space embeddings of $q(\theta)$, $q_i(z_i)$, as well as the equations (\ref{up1}) and (\ref{up2}). By definition given in equation(\ref{embed_eqn}),
\begin{equation}
\mu_\theta = \int_\theta \phi(\theta)q(\theta) d\theta \;\quad \quad 
\mu_{z_i} = \int_{z_i} \phi(z_i)q_i(z_i) dz_i
\end{equation}
The variational update equations in (\ref{up1}) and (\ref{up2}) 
provide us with the key relationships
between latent variables in the model. 
We can replace the specific distributional forms in these equations
with operators that maintain the same relationships 
among distributions represented in the Hilbert space embeddings.
%
\begin{equation}\label{complex}
q(\theta) = f_1(\theta, \{q_i(z_i)\}) \;\quad
q_i(z_i) = f_2(z_i, w_i, q(\theta))
\end{equation}

Here, $f_1$ and $f_2$ represent the abstract structure of the model implied by (\ref{up1}) and (\ref{up2}) without specific distributional forms. 
We will provide a specific instantiation of $f_1$ and $f_2$ shortly.
Following the same argument as in equation (\ref{hil1}), we can write equation (\ref{complex}) as $q(\theta) = \tilde{f_1}(\theta, \{\mu_{z_i}\})$. Similarly, $q_i(z_i) = \tilde{f_2}(z_i, w_i, \mu_\theta)
$. Iterating through all values of $\theta, z_i$ and using the operator view given in equation (\ref{hil2}) as reference, we get the following equivalent fixed-point equations in the Hilbert Space:  

\begin{equation}
   \mu_{\theta} = \mathcal{T}_1  \circ   \{ \mu_{z_i} \} \;\qquad \qquad
   \mu_{z_i}    = \mathcal{T}_2  \circ  [ w_i, \mu_{\theta}  ]
\end{equation}
\subsection{Parameterization of Hilbert space embedding using Deep Neural Networks}

The operators $\mathcal{T}_1$ and $\mathcal{T}_2$ 
have complex non-linear dependencies on the unknown true probability distributions 
and the feature map $\phi$. 
Thus, we need to model these operators 
in such a way that we can utilize the available data 
to learn the underlying non-linear functions. 
We will use deep neural networks 
which are known for their ability to model non-linear functions.

We start by parameterizing the embeddings. 
We assume that any point in the Hilbert space is a vector $\mu_i\in {\rm I\!R^{D}}$. 
Next, as the operators are non-linear function maps, we replace them by deep neural networks. 
In its simplest form, we only use a single fully connected layer 
with `$\tanh$' activations yielding the following fixed point update equations,

\begin{eqnarray}\label{param update}
   \mu_{\theta} = \tanh(\; W_1 \cdot \sum_{i=1}^{N}  \{\mu_{z_i}\}  \;)   \\
   \mu_{z_i} = \tanh(\; W_2 \cdot word2vec(w_i) + W_3 .\mu_{\theta}  \;)   \label{param 2}
\end{eqnarray}

The original work on Hilbert space embeddings required the embeddings to be injective. We observe that we do not need the embedding to be injective on the domain of all distributions. Instead, we only need it to be injective on the sub-domain of distributions used in the training corpus. The supervised training process on the training set will have to find embeddings that allow the model to distinguish documents that occur in the corpus automatically causing the learned embeddings to be injective for the training domain.

We keep the dimension of the $word2vec$ \cite{mikolov2013distributed} embedding identical to the Hilbert space embedding, i.e. $w_i \in {\rm I\!R^{D}}$.
Note, that the above parameterization is one example. 
Multiple fully connected layers can be used to achieve denser models. 

Assume the parameters $word2vec$, $W_1$, $W_2$ and $W_3$ are known. We calculate the set of embeddings for a given text corpus 
by iterating equations(\ref{param update}, \ref{param 2}). 
Algorithm~\ref{Hilbert embedding} summarizes this procedure. We normalize the embeddings after every iteration to avoid overflow. This is the heart of the 
Cooperative Neural Network paradigm in which a set of neural networks
co-constrain each other to produce an embedding informed by prior structure. In our experience, we found that `$\tanh$' works better than `$\sigma$' as a choice for non-linearity. Using rectified linear `ReLU' units will not work as they zero out negative values of the embeddings. We apply dropout 
\cite{JMLR:v15:srivastava14a} 
to $\mu_{z_i}$'s, $\mu_\theta$ and \textit{word2vec} for regularization. For every document, the algorithm returns the associated $\mu_\theta$ embedding, representing the document in the Hilbert space.

\begin{minipage}[t]{0.45\textwidth}
\vspace{0pt}
\begin{algorithm}[H]
   \caption{Getting Hilbert Space Embeddings}
   \label{Hilbert embedding}
\begin{algorithmic}
   \STATE {\bfseries Input:} Parameters $\{W_1, W_2, W_3\}$
   \STATE Initialize $\{\mu_{\theta}^{(0)}, \mu_{z_i}^{(0)}\} = \bf{0} \in {\rm I\!R^{D}}$.
   \FOR{$t=1$ {\bfseries to} \text{T iterations}}
     \FOR{$i=1$ {\bfseries to} $N$ words} 
     \STATE $\mu_{z_i}^{(t)} = \tanh(W_2 . word2vec(w_i) + W_3 .\mu_{\theta}^{(t)})$
     \STATE \text{Normalize } $\mu_{z_i}^{(t)}$
     \ENDFOR
      \STATE $\mu_{\theta}^{(t)} = \tanh(W_1 . \sum_{i=1}^{N} \{\mu_{z_i}^{(t-1)}\})$
      \STATE \text{Normalize } $\mu_{\theta}^{(t)}$
   \ENDFOR 
   \STATE return $\{\mu_{\theta}^{(T)}\} : \text{Document embeddings}$
\end{algorithmic}
\end{algorithm}
\end{minipage}%
\qquad
\begin{minipage}[t]{0.50\textwidth}
\vspace{0pt} 
\begin{algorithm}[H]
   \caption{Training using Hilbert Space Embeddings}
   \label{training Hilbert}
\begin{algorithmic}
   \STATE {\bfseries Input:} Document Corpus $\mathcal{D}$, with each doc `$d$' has set of words $[w_{d,i}] \in N_d$.
   \STATE \textit{Initialize} $\bf{P^{(0)}} = \{\bf{W}^{(0)}, \bf{u}^{(0)}, {word2vec}^{(0)}\}$ with random values. Let `learning rate = $r$'.
	\FOR{$t=1$ {\bfseries to} $\mathcal{T}$}
   \STATE Sample docs from $\mathcal{D}$ as $\{D_s, y_{s}\}$
   \STATE Using Alg(\ref{Hilbert embedding}) get Hilbert embeddings \{$\mu_{\theta_d}^{s}$\} for `$D_s$'
    \STATE $y_{pred} = \mathcal{H}\left(\mu_{\theta_d}^{s};\bf{P^{(t-1)}}\right)$
    \STATE Update: $\bf{P^{(t)}}$ = $\bf{P^{(t-1)}}$ - $r. \bigtriangledown_{\bf{P^{(t-1)}}} L(y_{pred}, y_s)$
	\ENDFOR 
   \STATE return $\{\bf{P^{\mathcal{T}}}\}$
\end{algorithmic}
\end{algorithm}
\vskip 0.1in
\end{minipage}%

In practice, the parameters $word2vec$, $W_1$, $W_2$ and $W_3$ are not known and need to be learned from training data. This requires formulating an objective function, and then optimizing that objective function. An additional advantage of the proposed method is that it allows using a wide variety of objective functions. In our case, we trained the model using a discriminative/supervised criterion that relies on the labels associated with each document, and we used binary cross-entropy loss or cross-entropy loss for multiclass classification. 

Algorithm~\ref{training Hilbert} summarizes the training procedure. It uses Algorithm~\ref{Hilbert embedding} as a subroutine. 
The $\mathcal{H}$ function is chosen to be a single fully connected layer in our implementation, which transforms the input embedding to a vector corresponding to number of classes. We sample (without replacement) a batch of documents $D_s$ from the corpus, compute their embeddings and update the parameters. The loss function takes in the $\mu_{\theta}$ embeddings and the corresponding document labels. The resulting model, called `CoNN-sLDA' is schematically illustrated in Figure~\ref{FigEmbeddingIllustration}. 

The CoNN-sLDA model retains the overall structure of the LDA model 
by separating the problem into document topic distributions 
and word topic distributions within each document. 
As with traditional LDA, one can visualize a document corpus 
by projecting topic vectors associated with documents 
into a 2D plane (e.g., using MDS, tSNE). 
An advantage of CoNN-sLDA over typical neural network approaches is that typical DNNs produce only a single embedding, whereas CoNN-sLDA elegantly factors 
the local and global information into separate parts of the model.
An advantage of CoNN-sLDA over traditional probabilistic graphical models is that
we can use low-bias, highly expressive distributions
implied by the neural network implementations of update operators.

\section{Experiments}

\subsection{Description of Datasets}
We evaluated our model `CoNN-sLDA' on two real-world datasets. The first dataset is a multi-domain sentiment dataset (MultiSent) \cite{blitzer2007biographies}, consisting of 342,104 Amazon product reviews on 25 different types of products (apparels, books, DVDs, kitchen appliances, $\cdots$). 
For each review, we go through the ratings given by the customer (between $1$ to $5$ stars) and label a it as positive, 
if the rating is higher than 3 stars and negative otherwise. 
We pose this as a binary classification problem. 
The average length of reviews is roughly 210 words after preprocessing the data. 
The ratio of positive to negative reviews is $\sim 8:1$. 
We use 5-fold cross validation
and report the average area under the ROC curve (AUC), in \%. 

The second dataset is the 20 Newsgroup dataset\footnote{ http://qwone.com/~jason/20Newsgroups/}. It  has around 19,000 news articles, divided roughly equally into 20 different categories. We pose this as a multiclass classification problem and report accuracy over 20 classes. The dataset is divided into training set (11,314 articles) and test set (7,531 articles), approximately maintaining the relative ratio of articles of different categories. The average length of documents after preprocessing is $\sim 160$ words. This task becomes challenging as there are some categories that are highly similar, making their separation difficult. For example, the categories ``PC hardware'' and ``Mac hardware'' have quite a lot in common.

We apply standard text preprocessing steps to both datasets. We convert everything to lower case characters and remove the standard stopwords defined in the `Natural Language Toolkit' library. We remove punctuations, followed by lemmatization and stemming to further clean the data. However, for other classifiers, we use the preprocessing techniques recommended by the respective authors. 

\subsection{Baselines for comparison}

We compare `CoNN-sLDA' with existing state-of-the-art algorithms for document classification.
We compare against VI-sLDA, \cite{chong2009simultaneous,mcauliffe2008supervised}, which includes the label of the document in the graphical model formulation and then maximizes the variational lower bound. Different from VI-sLDA, the supervised topic model using DiscLDA \cite{lacoste2009disclda} reduces the dimensionality of topic vectors $\theta$ for classification by introducing a class-dependent linear transformation. 

Boltzmann Machines are traditionally used to model distributions and with the recent development of deep learning techniques, these approaches have gained momentum. We compare with one such Deep Boltzmann Machine developed for modeling documents called Over-Replicated Softmax (OverRep-S) \cite{srivastava2013modeling}. Another popular approach is by \cite{chen2015end}, called BP-sLDA, which does end-to-end learning of LDA by mirror-descent back propagation over a deep architecture.
We also compare with a recent deep learning model developed by \cite{chien2018deep} called DUI-sLDA.


\subsection{Classification Results}


Table(\ref{table-newsgroup}) shows the accuracy results on newsgroup dataset together with standard error on the mean (SEM) over 5 folds. For each of 5 folds, we split training data into train and validation and optimize all parameters. We then evaluate against a fixed common test set. As the number of classes is 20, we found that using higher Hilbert space dimensions work better (See entries for Dim=40 and Dim=80 in table). A dropout of $\sim 0.8$ was applied to \textit{word2vec} embeddings. The batch size was fixed at 100 and we trained for around 400 batches. The performance of CoNN-sLDA is better than BP-sLDA and at par with 5 layer DUI-sLDA model. The cost sensitive version CoNN-sLDA (Imb), balances out the misclassification cost for different classes in the loss function tends to perform slightly better. 
The 20 newsgroup dataset is one of the earliest and most studied text corpuses.
It is fairly separable, so most modern state-of-the-art methods do well on it, 
but it is an important benchmark to establish the credibility of an algorithm.

Our CoNN-sLDA model was able to outperform the recently proposed state-of-the-art method, DUI-sLDA, on the large `MultiSent' dataset (table\ref{table-MultiSent}) having over 300K documents by a significant AUC margin of 2\%. This corresponds to a 23\% reduction in error rate. 
We used a single fully connected layer with $\tanh$ non-linear function for both, $\mu_\theta, \mu_{z_i}$ embeddings. 
Hilbert space dimension and \textit{word2vec} dimension are both $10$. We use a dropout probability of $0.1$, 
The Algorithm(\ref{Hilbert embedding}) was unrolled for $1$ iteration. `Batch size' was set at $100$ and ran for $3000$ batches with optimization done using `Adam' optimizer. 
We also ran a cost sensitive version of CoNN-sLDA (Imb) model, with a balancing ratio of 1.4 towards the minority class which was incorporated in the loss function. We observe slight improvement in results. CoNN-sLDA consistently outperformed other models over various choices of model parameters, see Appendix(B).


\begin{table}
%
  
  \parbox[t]{.45\linewidth}{  
    \centering
    \begin{small}
      \begin{tabular}{lll}
        \toprule
        Classifier & Accuracy(\%) & Details \\
        \midrule
        VI-sLDA & 73.8$\pm$ 0.49& $K\seq50$\\ 
        DiscLDA & 80.2$\pm$ 0.45& $K\seq50$ \\ 
        OverRep-S & 69.5$\pm$ 0.36& $K\seq512$ \\ 
        BP-sLDA & 81.8$\pm$ 0.36& $K\seq50,L\seq5$\\ 
        DUI-sLDA& 83.5$\pm$ 0.22& $K\seq50,L\seq5$\\ 
        CoNN-sLDA & 83.4 $\pm$ 0.18 & $\textrm{Dim}\seq40$\\ 
        CoNN-sLDA(imb) & \textbf{83.7$\pm$ 0.13}& $\textrm{Dim}\seq80$\\ 
        \bottomrule
      \end{tabular}
      \caption{{
         `20 Newsgroups' classification accuracy on 19K documents. 
          SEM over 5 fold CV. Dim indicates Hilbert space dimension.}}
      \label{table-newsgroup}
    \end{small}
  }
  \hspace{0.25in}
  \parbox[t]{.45\linewidth}{
    \centering
    \begin{small}
      \begin{tabular}{lll}
        \toprule
        Classifier & AUC (\%) & Details \\
        \midrule
        VI-sLDA    & 76.8$\pm$ 0.40& $K\seq50$ (topics)\\ 
        DiscLDA    & 82.1$\pm$ 0.40& $K\seq50$ \\ 
        BP-sLDA    & 88.9$\pm$ 0.36& $K\seq50,L\seq5$\\ 
        DUI-sLDA     & 86.0$\pm$ 0.31& $K\seq50,L\seq1$\\
        DUI-sLDA     & 91.4$\pm$ 0.27& $K\seq50,L\seq5$\\
        CoNN-sLDA     & \textbf{93.3$\pm$ 0.13}& $\textrm{Dim}\seq10$\\ 
        CoNN-sLDA(imb)& \textbf{93.4$\pm$ 0.13}& $\textrm{Dim}\seq20$\\ 
        \bottomrule
      \end{tabular}
      \caption{{`MultiSent' AUC on 324K documents. 
               SEM over 5 Fold CV. Dim indicates Hilbert space dimension.}}
      \label{table-MultiSent}
    \end{small}
  }
  \vskip -0.2in
\end{table}

The number of layers required by other deep models like DUI-sLDA, BP-sLDA for good classification is usually quite high and their performance decreases considerably with fewer layers. CoNN-sLDA outperforms them with a single layer neural network.

We have a vectorized and efficient implementation of CoNN-sLDA in PyTorch and Tensorflow. The results shown above are from the PyTorch version. We ran our experiments on NVIDIA Tesla P100 GPUs. The runtime for 1 fold of `MultiSent' for the settings mentioned above is around \textit{5 minutes}, while a single fold for `20 Newsgroup' dataset runs within \textit{2 minutes}.

In Appendix(B), we report our experiments to optimize the algorithmic and architectural hyperparameters. We use the `MultiSent' data for our analysis. In general for training, we recommend starting with a small Hilbert space dimension and batch size, then try increasing the number of fully connected layers and finally choose to unroll the model further. 

\begin{figure}[ht]
  \begin{center}
  \centerline{
     \includegraphics[width=0.8\columnwidth]
          {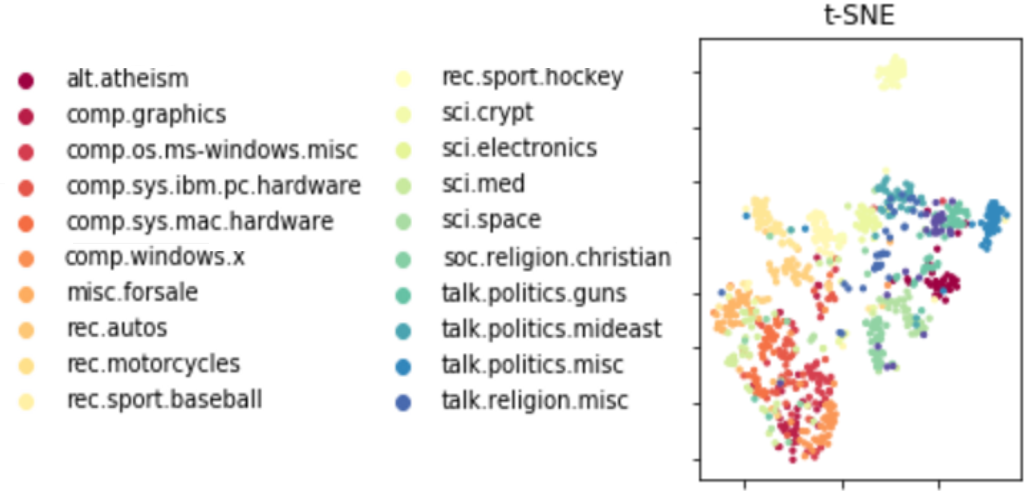}} 
  \caption{A t-SNE projection of the 40-dimensional
     embeddings $\mu_\theta$ for test documents in the 20-Newsgroups 
     dataset. The colors represent the category label for each 
     document. The embeddings separate categories very well.}
  \label{tsne_viz_newsgroups}
  \end{center}
\end{figure}

\section{Discussions \& Future extensions}

\begin{figure}
  \qquad \qquad
  \begin{minipage}[c]{0.5\textwidth}
    \caption{
       tSNE visualization of a random sample of 10-dimensional $\mu_\theta$
      embeddings for Multisent documents (Blue positive, red negative). 
       The embeddings project distinct categories to highly coherent regions.
    }\label{tsne_viz_multisent}
  \end{minipage}
  \begin{minipage}[c]{0.9\textwidth}
    \includegraphics[trim={18cm 0 0 0.75cm},clip,width=0.3\columnwidth]{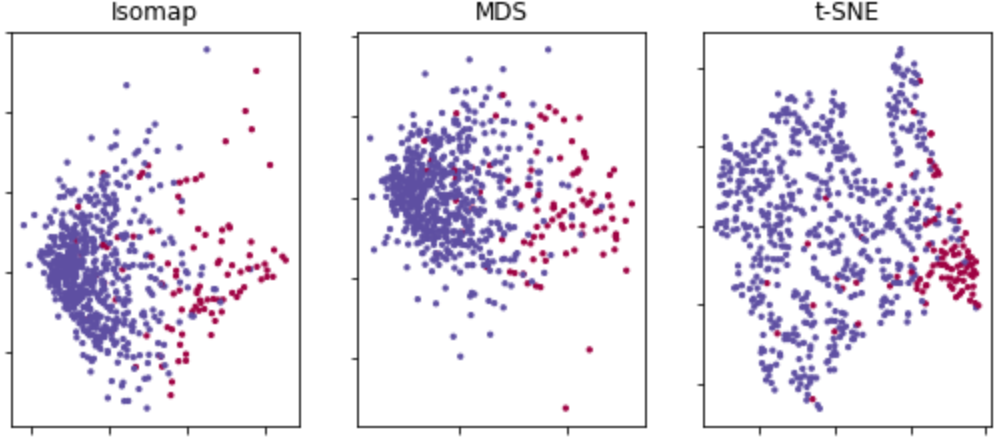}
  \end{minipage}\hfill
\end{figure}

In addition to supervised classification, we can use LDA style models
for visualizing and interpreting the cluster structure of the datasets.
For example, in CoNN-sLDA model, we can use t-SNE 
\cite{maaten2008visualizing} 
to visualize the documents using their $\mu_\theta$ values. In Figure~\ref{tsne_viz_newsgroups} we see that CoNN-sLDA clearly maps
different newsgroups to homogeneous regions of space
that help classification accuracy and 
provide insight into the structure of the domain.
Similarly, Figure~\ref{tsne_viz_multisent} shows that CoNN-sLDA
maps the positive and negative product reviews into 
different regions facilitating classification and interpretation.

An interesting extension for the CoNN-sLDA model will be to map the Hilbert space topic embedding $\mu_\theta$ back to the original topic space distribution. This would potentially allow us to provide text labels for the discovered clusters providing an intuitive interpretation for the model learned by our technique. Appendix (C) discusses an approach to get most relevant words in a document pertaining to a discriminative task.

In this work, we obtain the fixed point update equations using 
the mean-field inference technique. In general, we can extend this procedure to other variational inference techniques. For example, we can find embeddings for Algorithm~\ref{Hilbert embedding} by minimizing the free energies of loopy belief propagation or its variants (e.g., \cite{wainwright2003tree}) and use Algorithm~\ref{training Hilbert} to train them end-to-end.

%

\section{Conclusion}

Cooperative neural networks (CoNN) 
are a new theoretical approach 
for implementing learning systems which can 
exploit both prior insights about the independence structure of the problem domain
and the universal approximation capability of deep networks.
We make the theory concrete with an example, CoNN-sLDA, 
which has superior performance to both 
prior work based on the probabilistic graphical model LDA and 
generic deep networks.
While we demonstrated the method on text classification 
using the structure of LDA, 
the approach provides a fully general methodology 
for computing factored embeddings 
using a set of highly expressive networks.
Cooperative neural networks thus expand the design space 
of deep learning machines in new and promising ways.

\section*{Acknowledgements}

We are thankful to our colleagues Srinivas Eswar, Patrick Flick and Rahul Nihalani for their careful reading of our submission.
%
{
\small
\bibliography{references}
\bibliographystyle{plain}
}
\clearpage

\section*{Appendix}

\subsection*{A. Derivation of fixed point equations}\label{fixed_point}

Inference in LDA requires estimating the distribution over $\theta$ and $\textbf{z}$.
Using the Bayes rule, this posterior can be written as follows: 
\begin{equation}
   p( \theta, \textbf{z} | \textbf{w}, \alpha, \beta ) = 
             \frac{  p( \theta, z, w | \alpha, \beta ) }
                  {  p( w | \alpha, \beta ) }
\end{equation}

To perform variational approximation, 
we approximate this LDA posterior with the PGM as shown in Figure~\ref{lda_pgm2}.


The joint distribution for the approximate PGM is given by:
\begin{equation}
%
   q( \theta, z ) = 
         q( \theta ) { \displaystyle \prod_{i=1}^{N} q_i( z_{i} )  }
\end{equation}

We want to tune the approximate distribution to resemble the true posterior as much as possible. 
To this end, we minimize the KL divergence 
between the two distributions. 
Alternatively, this can be seen as minimizing the variational free energy 
of the Mean-Field inference algorithm \cite{wainwright2008graphical}: 
\begin{equation}
   \underset{ \{q\} }{\rm min} 
         \left\{ D_{KL} (\;          q( \theta, z ) 
                            \;||\;   p( \theta, z  |  w, \alpha, \beta  )   \;)  \right\}
\end{equation}
Substituting the expression for KL-divergence, we get
\begin{equation}
   \underset{  \{q\} }{\rm min} 
           \int_{\theta} \idotsint_{\{z_i\}} 
                     q( \theta, z )  \;
                     \log{  \frac{ q( \theta, z ) } 
                                 { p( \theta, z | w, \alpha, \beta ) }  } 
           \; d\theta \;  \{ {dz}_{i} \}
\end{equation}

Using the Bayes formulation given in equation(\ref{Bayes}) 
and observing that $p(w| \alpha, \beta)$ is a constant, 
we can write
\begin{equation}
  \begin{split}
     \underset{ \{q\} } {\rm min} 
               \int_{\theta}\idotsint_{\{z_i\}} 
                       q(\theta, z) \;
                       [\;\;  \log{ q(\theta, z) }  
                            - \log{ p( \theta, z, w | \alpha, \beta ) }  \;\;] 
              \;d\theta \; \{{dz}_{i}\}
  \end{split}
\end{equation}
Substituting the probability densities given in equations (\ref{true}) and (\ref{e2}), 
the following minimization expression is obtained:
\begin{equation}
   \begin{split}
      \underset{ \{q\} } {\rm min}
             &  \int_{\theta} \int_{\{z_i\}} 
                    \left\{   q(\theta)    { \displaystyle \prod_{i=1}^{N} q_i( z_{i} ) }   \right\}
                    \left\{\;
                        \log\left(  q( \theta ) 
                                    { \displaystyle \prod_{i=1}^{N} q_i( z_{i} )  }   \right)   \right.\\
                    & \left. 
                        - \log \left(  
                                 p( \theta | \alpha ) 
                                 { \displaystyle \prod_{i=1}^{N} p( z_i | \theta ) 
                                                                 p( w_i | z_i, \beta ) } 
                             \right)
                    \; \right\}  
               \; d\theta \; \{{dz}_{i}\}
\end{split}
\end{equation}


Pulling logarithms inwards we can convert products to summations.
We then move integrals inward. 
In some cases, integrals add up to 1 
( {\em e.g.},  $\int_{\theta} q(\theta)\; \; d\theta = 1$).
In some cases, inner sums can be pulled outwards.
The result consists of simple integrals:
\begin{equation} \label{e9}
   \begin{split}
      \underset{ \{q\} } {\rm min } \Big \lbrace \;\;
              & \int_{\theta}  q(\theta) \log{ q(\theta) }  \; d\theta 
                  + { \displaystyle \sum_{i=1}^{N} \int_{z_i} q_i(z_i)\log{q_i(z_{i})} } \; {dz}_{i} 
                 - \int_{\theta} q(\theta)\log{p(\theta|\alpha)} \; d\theta\\  
                 &- {\displaystyle \sum_{i=1}^{N} \iint_{\theta, z_i} 
                                                            q(\theta) q_i(z_i) 
                                                            \log{p(z_i|\theta)}} 
                                                        \; d\theta \; {dz}_{i} 
                 - {\displaystyle\sum_{i=1}^{N}  \int_{z_i} q_i(z_i)\log{p(w_i|z_i, \beta)} } \; {dz}_{i}
          \;\; \Big \rbrace
\end{split}
\end{equation}

We denote the expression given in equation(\ref{e9}) by 
$\underset{ \{q\} }{ \rm min }(L)$. 
To minimize the functional equation given by $L$, 
we take the functional derivatives of $L$ 
with respect to $q(\theta)$ and $q_i(z_i)$ 
and equate them to zero.


Solving for
$\left(\frac{\delta L}{\delta q(\theta)} = 0\right)$, 
we get the first fixed point equation:
%
\begin{equation}
   \begin{split}
       \log{q(\theta)} 
               =  \log{p(\theta|\alpha)} 
                  +  {\displaystyle \sum_{i=1}^{N} 
                                \int_{z_i} q_i(z_i) \log{p(z_i|\theta)}} 
                                     \;dz_{i} 
                  - 1
   \end{split}
\end{equation}

Similarly, solving for $\frac{\delta L}{\delta q_i(z_i)} = 0$, we get the second fixed point equation:
%

\begin{equation}
   \begin{split}
      \log{q_i(z_{i})} =  \log{p(w_i|z_i, \beta)} 
                          + \int_{\theta} q(\theta) \log{p(z_i|\theta)} d\theta -1
  \end{split}
\end{equation}


Note that this derivation is different from the classical variational approximation derivations, where the EM algorithm is eventually used to iteratively approximate the posterior.

\subsection*{B. Architecture choices of  CoNN-sLDA model}
In this section we report on our experiments to optimize the
algorithmic and architectural hyperparameters.
We use the `MultiSent' dataset for our analysis.
\subsubsection*{B.1 Varying Hilbert Space Embeddings dimension}
\begin{figure}[ht]
\begin{center}
\centerline{\includegraphics[width=0.6\columnwidth]{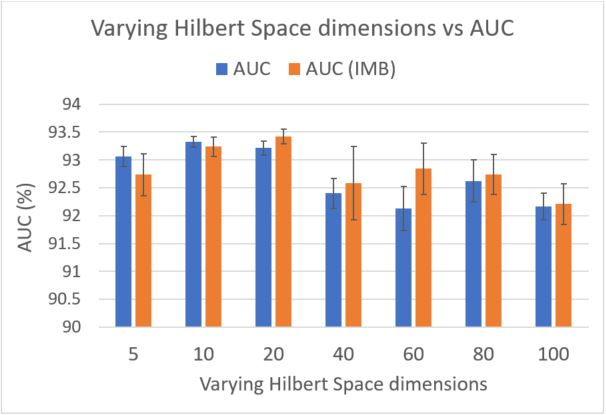}}
\caption{Varying Hilbert space embeddings dimension along the x-axis and the AUC values on y-axis. We also compare with the cost sensitive learning version, denoted by AUC (IMB) for every dimension choice. The depth of neural networks for both the embeddings $\mu_\theta$ and $\mu_{z_i}$ is a single fully connected layer. }
\label{vary-hilbert}
\end{center}
\vskip -0.2in
\end{figure}

The dimensionality of the Hilbert space trades off the expressive power against computation and storage requirements of the model.
In Figure~\ref{vary-hilbert}, we show varying Hilbert space dimensions on the x-axis and compare their AUCs. We observe a decline in AUC after Hilbert dimension of $20$. We postulate that higher Hilbert space dimensions tend to overfit the data. 
Empirically we found that with lower Hilbert space dimensions we have to scale down the dropout appropriately.

As the data is imbalanced between number of positive and negative reviews, we did cost sensitive learning in CoNN-sLDA (Imb) by adjusting the weights of the loss function for different classes and were able to attain slight improvement.  

\subsubsection*{B.2 Varying number of Iterations of update equations in 
Algorithm(\ref{Hilbert embedding})}
Figure(\ref{vary-itr}) shows the plot of varying number of iterations of update equations in algorithm(\ref{Hilbert embedding}) versus the AUC obtained. We can observe the that AUC decreases and the corresponding standard deviation increases as we increase the number of iterations.
In our experience, our algorithm works well even for a single iteration and going beyond 5 iteration gives no significant improvement in results. 


\subsubsection*{B.3 Varying depth of the model}

\begin{figure}[h]
\centering
\begin{subfigure}{.37\textwidth}
  \centering
  \includegraphics[width=.9\linewidth]{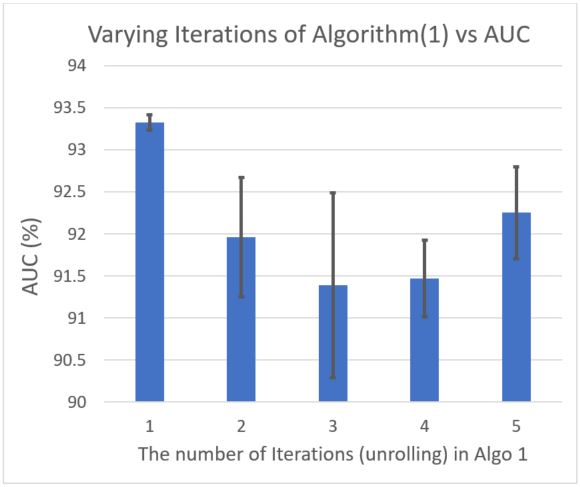}
  \caption{Varying iterations}
  \label{vary-itr}
\end{subfigure}%
\begin{subfigure}{.5\textwidth}
  \centering
  \includegraphics[width=0.9\linewidth]{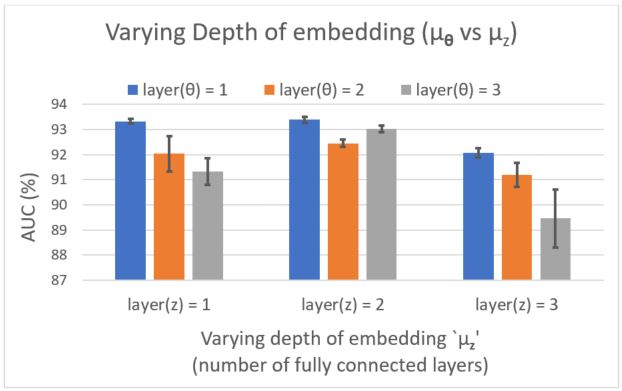}
  \caption{Varying $\mu_\theta$ and $\mu_{z_i}$'s}
  \label{vary-depth}
\end{subfigure}
\caption{(a) Unrolling the model along the x-axis and AUC on the y-axis. For the `MultiSent' dataset, we found that even using a single iteration works well. (b) Plot showing number of fully connected layers for various combinations of $\mu_\theta$ and $\mu_{z_i}$'s. The AUC values are shown on y-axis. We observe that the setting where there are two fully connected layer of embedding $\mu_{z_i}$ consistently gives good results for varying layers of embedding $\mu_\theta$. }
\label{combined_expt}
\end{figure}

In Algorithm~\ref{Hilbert embedding}, we parameterized the embeddings $\mu_\theta$ and $\mu_{z_i}$ using deep neural networks. Here, we analyze the results of varying the depth of the neural networks and their effect on the corresponding AUC.
Figure~\ref{vary-depth} shows a combination plot, where we visualize the AUC values for various different combinations of depth between $\mu_\theta$ and $\mu_{z_i}$. 


We found that two fully connected layers for embedding $\mu_{z_i}$ and a single fully connected layer for embedding $\mu_\theta$ works well for both datasets. 
Deeper models tend to overfit the data.
For training, we recommend starting with a small Hilbert space dimension and batch size, then increase the number of fully connected layers, and finally choose to unroll the model further. 

\subsection*{C. Interpretability: Getting the relevant words based on embeddings obtained}

The embedding model defines a relationship between words found in documents $w$
and topic distributions for documents $\mu_{\theta}$.
Usually we calculate the topic distribution from the words in documents.
For interpretation purposes, we might wish to go the other direction:
from topics $\mu_{\theta}$ to words in the topic. 
For instance, after running CoNN-sLDA, we get embeddings for all of the 
documents. We could then cluster these to get $K$ clusters.
We might then ask how to interpret these clusters.
We could take the mean embedding of each cluster $\mu_k$
and recover the words that would be associated with the cluster
(e.g., SLR, aperture, resolution versus click-and-shoot, special effects).
Alternatively, we could run PCA on the embedding space to find 
the principle directions of variation of document topics.
We can then recover the words associated with the end-points of
each distribution in order to label this dimension
(e.g., light weight versus heavy or easy-to-use versus complicated).

We show here how to define a relationship between a given $\mu_{\theta}$
and the words associated with the topic. 
Given the $\mu_{\theta}$ from CoNN-sLDA model of a document under consideration, we want to find the top $word2vec$ vectors which satisfies the equation(\ref{inter}). If we substitute \ref{param 2} into
\ref{param update}, we can eliminate the dependence on 
word topic distributions $z_i$. 
\begin{equation}\label{inter}
   \mu_{\theta} = \tanh(\;      W_1   
                          \cdot \sum_{i=1}^{N}  
                                  \{\tanh(\;       W_2 
                                             \cdot word2vec(w_i) 
                                             + W_3 .\mu_{\theta}  \;) \}  \;) 
\end{equation}
The $\mu_{\theta}$ terms are related
by the sum of the embeddings of words in the text.
The embeddings for the same word are always the same, 
so we can group all embeddings for word class $c$ together
and just keep a class weights $F_c$.
We set \label{inter2} to zero so that we have an equation that measures discrepancy between current system and a consistent system. 
We then form an objective $J_w$ which is a function of 
topic distribution $\mu_{\theta}$ and $K$ word class weights $F_c$.
\begin{equation}\label{inter2}
\begin{split}
 J_w(F_c; \mu_{\theta}) = \tanh(\; W_1 \cdot 
        \sum_{c=1}^{K} 
              F_c \{\tanh(\; W_2 \cdot word2vec(w_c)+ W_3 .\mu_{\theta}  \;)\}  \;) - \mu_{\theta}
\end{split}
\end{equation}
Minimizing the square of $J_w$ w.r.t. the $F_c$ parameter will give us the weights of the words relevant to the embeddings $\mu_\theta$. 
\begin{equation}
F_c^* = argmin_{F_c} J_w^2(F_c; \mu_\theta) 
\end{equation}
We can thus find the top most commonly occurring highly weighted words corresponding to any documents distribution embedding $\mu_{\theta}$
or examine words associated with any $\mu_{\theta}$ in the 
embedded space.







\end{document}